\documentclass{ifacconf}

\usepackage{graphicx}      
\usepackage{natbib}        
\usepackage{caption}
\usepackage{subcaption}
\usepackage{multirow}
\usepackage{tabularx}
\usepackage{rotating}
\usepackage{adjustbox}
\usepackage{float}
\usepackage{amsmath}
\usepackage{amsfonts}

\begin{document}
\begin{frontmatter}

\title{Transformer-based Capacity Prediction for Lithium-ion Batteries with Data Augmentation} 


\author[First]{Gift Modekwe} 
\author[First]{Saif Al-Wahaibi}
\author[First]{Qiugang Lu}\thanks{Corresponding author: Qiugang Lu; Email: jay.lu@ttu.edu}

\address[First]{Department of Chemical Engineering, Texas Tech University, Lubbock, TX 79409, USA} 

\begin{abstract}                
Lithium-ion batteries are pivotal to technological advancements in transportation, electronics, and clean energy storage. The optimal operation and safety of these batteries require proper and reliable estimation of battery capacities to monitor the state of health. Current methods for estimating the capacities fail to adequately account for long-term temporal dependencies of key variables (e.g., voltage, current, and temperature) associated with battery aging and degradation. In this study, we explore the usage of transformer networks to enhance the estimation of battery capacity. We develop a transformer-based battery capacity prediction model that accounts for both long-term and short-term patterns in battery data. Further, to tackle the data scarcity issue, data augmentation is used to increase the data size, which helps to improve the performance of the model. Our proposed method is validated with benchmark datasets. Simulation results show the effectiveness of data augmentation and the transformer network in improving the accuracy and robustness of battery capacity prediction.
\end{abstract}


\end{frontmatter}

\section{Introduction}
Lithium-ion batteries have become increasingly popular across various sectors due to their numerous benefits, including a longer lifespan, higher specific weight and energy density, faster charging, and a lower self-discharge rate compared to other types of batteries \citep{horiba2014lithium}. Their widespread adoption is driven by the growing demand for portable electronics \citep{morita2021estimation} and a shift towards more sustainable transportation \citep{miao2019current}. For the safety and optimal performance of these batteries, battery management systems are usually put in place to monitor the state of health (SOH). The SOH of a battery describes the current condition of the battery by comparing its present capacity to its nominal capacity \citep{yao2021review}. This is necessary because repeated charging, discharging, and internal chemical changes gradually diminish battery capacity, reducing the energy that it can deliver (i.e., lifetime) \citep{barcellona2022aging}. 
 
 Generally, capacity estimation is carried out by two main approaches: model-based approaches and data-driven approaches \citep{lucaferri2023review}. Model-based approaches use physics-based models to describe the gradual loss of capacity observed in lithium-ion batteries. These models aim to provide fundamental insights into how the capacity fade occurs with an attempt to offer physical explanations about the observed changes. Models that have been explored include the electrochemical models, equivalent circuit models, and semi-empirical models \citep{guo2022review}. A significant drawback of model-based approaches is their high computational complexity due to the large-scale coupled nonlinear partial differential equations (PDEs) involved, especially in the electrochemical models. 
 
 Alternatively, data-driven approaches directly utilize historical data for capacity prediction \citep{zhang2023review}. Examples include support vector machines, relevance vector machines, neural networks, and so on \citep{wu2016review}. Among others, neural networks (NNs) are usually preferred because they do not require manual feature engineering or additional processing that involves extensive expertise \citep{hannan2021deep}. Various types of NNs have been adopted for capacity prediction, including artificial neural networks \citep{hussein2014capacity}, convolutional neural networks (CNNs) \citep{lee2023convolutional}, recurrent neural networks \citep{catelani2021remaining}, and long short-term memory (LSTM) networks \citep{zhang2018long}. In particular, \cite{park2020lstm} examined the usage of LSTM by leveraging multi-channel battery profiles as a means of improving the remaining useful life prediction.  
 
 Despite the success achieved by these methods, capacity prediction of Lithium-ion batteries with recurrent networks often require extensive computation and may suffer from the inability to capture long-term dependencies in time-series data \citep{chen2022transformer}. In contrast, the transformer network possesses a significant advantage due to its self-attention and parallel processing ability, which makes it a good option for capturing long-range dependencies with computational ease \citep{vaswani2017attention}. Consequently, the transformer architecture has been applied to solve time series problems, including battery management (\cite{chen2022transformer}, \cite{gu2023novel}). 
 
 Meanwhile, a significant limitation of NN-based approaches is the need for extensive data. \cite{fan2022data} suggested a data augmentation method that involves adding Gaussian noise to the battery dataset to improve the flexibility and robustness of the model. One motivation is to use these random noises to mimic capacity regeneration, fluctuations, and sensor inaccuracies occurring during the usage and monitoring of batteries, thus enhancing the adaptability and resilience of the developed SOH estimation method. In this paper, we will leverage and integrate the benefits of transformers and data augmentation to improve the capacity prediction of Lithium-ion batteries. Benchmark NASA Group 1 and University of Michigan (UofM) battery aging datasets will be used to validate the effectiveness of the proposed method. 
 
 
\section{Preliminaries}
\label{sec: Preliminaries}

\subsection{Transformer Network}
The transformer network has drawn widespread attention in the natural language processing community \citep{wolf2020transformers}. It adopts the well-known attention mechanism, which captures the relationship between input elements in a sequence \citep{vaswani2017attention}. This mechanism enables the model to prioritize specific information, aiding in enhanced context comprehension and better handling of long-range relationships. The major parts of the transformer used in this study are discussed below.

\textit{Multi-head Attention (MHA):} MHA uses the attention mechanism to learn from and simultaneously attend to different representational subspaces of the input at various positions \citep{vaswani2017attention}. Due to its parallelizable nature, it captures more complex patterns and accelerates the training speed. Given an input sequence $X = \{x_1, x_2, …,x_n\}$, the attention mechanism projects the input elements into a subspace by using learned weight matrices $W_i^Q , W_i^V, W_i^K$, where $i$ refers to the $i$-th head in the attention:
\begin{equation}
Q_i = XW_i^Q, \quad K_i = XW_i^K, \quad V_i = XW_i^V,
\end{equation}
$Q$, $K$, and $V$ stand for the query, key, and value matrices, respectively. A scaled-dot product is done to obtain the attention scores that stabilize the gradients during training. Softmax normalizes the values obtained by the scaled-dot product, and the attention matrix $Z_i$ is thus obtained as \citep{vaswani2017attention}: 
\begin{equation}
Z_i= softmax\left(\frac{Q_i K_i^T}{\sqrt{d_k }}\right)V_i.
\end{equation}
For $m$ heads, the attention matrices are concatenated to form the multi-head attention matrix, which is then transformed back to the original input dimension by using a learnable weight matrix $W^O$:
\begin{equation}
Z=concat(Z_1,Z_2,…Z_m) W^O.
\end{equation}

\textit{Positional Encoding:} The scaled-dot product of the transformer architecture does not take into account the order of the input sequence. To address this issue, positional encoding is used to include the position information into the input embedding. The sinusoidal function is commonly adopted to compute the positional encoding, as shown below \citep{vaswani2017attention}:
\begin{equation}
PE(pos, 2j) = \sin\left(\frac{pos}{10000^{2j / d_{\text{model}}}}\right), \label{eq: posenc}
\end{equation}
where $PE(pos,2j)$ and $PE(pos,2j+1)$ represent the positional encoding for the even and odd index positions, respectively, $pos$ is the position of the element in the sequence, $j$ is the dimension ranging from 0 to $d_{model}/2$, and $d_{model}$ is the dimension of the embedding.
	
\textit{Feed-Forward Network:} The feed-forward neural network of the transformer is made up of an activated function, the Rectified Linear Unit (ReLU) and dense layers that separate two linear transformations. As a major part of the encoder, it is tasked with introducing non-linearity that can assist the transformer to learn intricate patterns for making better predictions. 
	
\textit{Encoder Block:} The transformer encoder consists of several identical blocks stacked on top of each other, with each block containing two layers (see Fig. \ref{fig3}). The first layer utilizes the multi-head attention for the positionally encoded input, while the second layer is a fully connected feed-forward neural network to process the extracted features from the attention mechanism \citep{vaswani2017attention}. Skip connections are utilized within each layer to mitigate vanishing gradient issues \citep{liu2021rethinking}, and layer normalization is also applied to enhance the model's stability and potentially speed up the training. Such a block is repeated several times to enhance the contextual understanding of the input sequence.

\subsection{Data Augmentation}
Data augmentation is commonly employed in machine learning (ML) to overcome the issue of data scarcity \citep{mumuni2022data}. It can be achieved through generating synthetic data and expanding existing dataset to increase the performance and robustness of ML models.

In this study, to address the data scarcity issue, data augmentation is done by adding Gaussian noise to the original dataset to obtain a variant of the input having a similar distribution to the original dataset. For simplicity, Gaussian Noise  \( G \sim \mathcal{N}(0,\sigma^2) \) is added to every individual sample to create a new dataset:
\begin{equation}
\bar{D} = D + E,
\end{equation}
where $\bar{D}$ is the new dataset, $D$ is the original dataset and $E$ is the noise obtained from sampling a Gaussian distribution with mean 0 and standard deviation $\sigma$. The zero mean ensures that the central tendency of the data remains unchanged after augmentation. However, the introduction of noise with varying $\sigma$ allows the model to accommodate potentially unseen variations in the data, which is particularly useful for handling real-world fluctuations in the measurements. Through this method, more variability is introduced and the training set is expanded. 



\section{Battery Dataset}
\label{sec: Battery Dataset}
To validate the proposed method, we use data from Group 1 of the NASA Ames Prognostic Center of Excellence Aging Dataset \citep{saha2007battery} and 4 cells from the dataset generated by UofM \citep{mohtat2021reversible}. The NASA Group 1 dataset consists of 4 cells labeled as B0005, B0006, B0007, and B0018. These cells were charged using a constant-current constant-voltage (CC-CV) protocol where the current was kept constant until the voltage reached 4.2V, and then the voltage was maintained while the current dropped to 20 mA. The experiment ended when the nominal voltage of the cells decreased to 2.7V, 2.5V, 2.2V, and 2.5V for the 4 cells, respectively. The experiment was carried out at an ambient temperature of 24$^\circ$C. For the  UofM dataset, the 4 cells considered were identified as M1, M4, M7, and M10. The cycling procedure includes charging the batteries with a CC-CV protocol similar to that of NASA, although in this case the current dropped to less than C/50 rate (i.e., fully charging the battery needs 50 hours). The experiment ends when the voltage decreased to 3.0V for the four cells and it was performed at room temperature (25$^\circ$C). Note that each of the 4 cells has a different C-rate for their charge and discharge process. Fig. \ref{fig1} below shows the capacity degradation over charging-discharging cycles for NASA Group 1 and UofM datasets, respectively.

\begin{figure}[!h]
	\centering
	\includegraphics[width = \columnwidth]{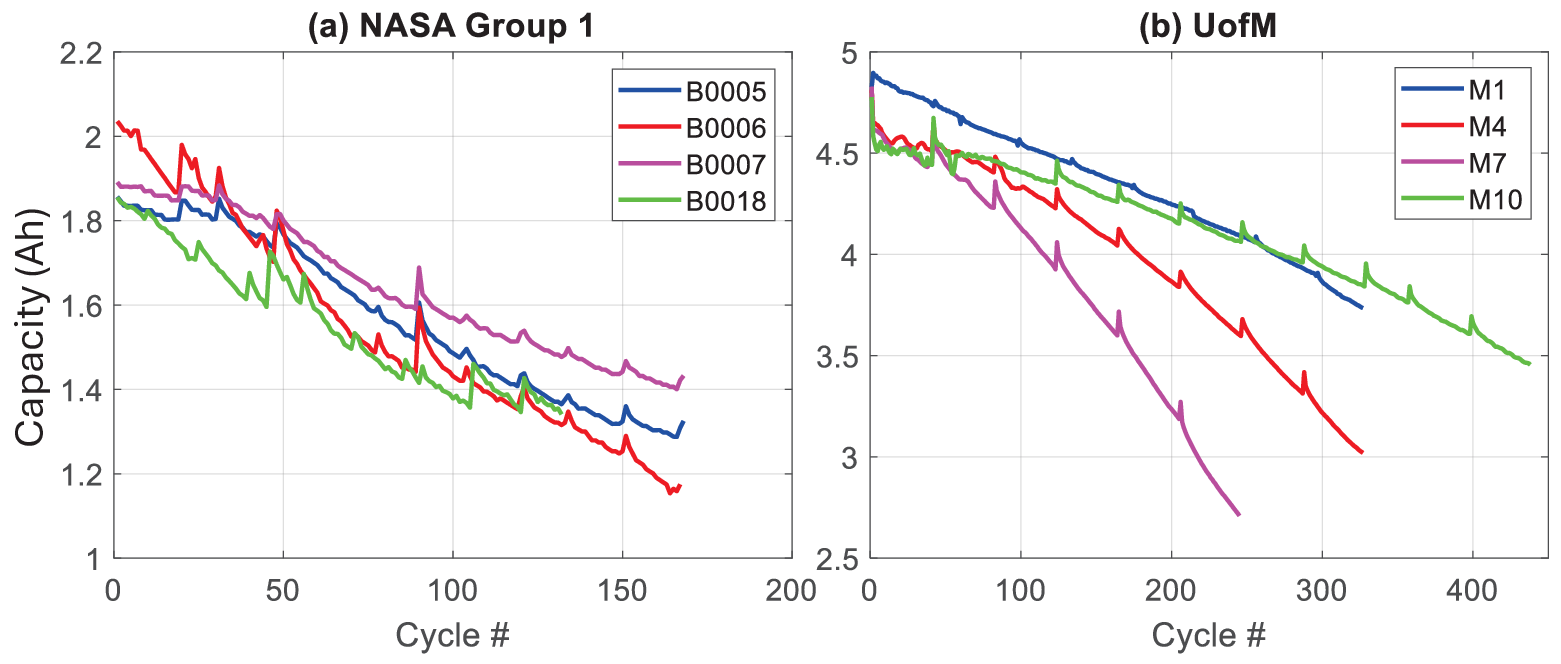}
	\vspace{-15 pt}
	\caption{Capacity degradation profile for the batteries used in this work.}
	\label{fig1}
\end{figure}

In this study, we focus on the charging cycles as they offer a more stable environment for measuring the internal battery parameters, which are otherwise challenging to gauge during discharging because of the swift and unpredictable current variations. Additionally, the discharge rates are often dictated by the battery owner's usage patterns, which introduces a high degree of variability \citep{choi2019machine}.   

\section{Transformer-based capacity prediction with multi-channel Profiles}
\label{sec: Transformer-based capacity estimation with multi-channel data}

\subsection{Multi-channel Profiles for Capacity Prediction}
This study integrates multi-channel battery profiles, including current, voltage, and temperature, for a charging cycle to capture the predictive relationship between such profiles and battery capacity values. Our motivation for using multi-channel profiles arises from the fact that as battery degrades, the patterns of charging profiles will change, which can serve as an indicator of the degradation status.  We adapt the transformer architecture to forecast future capacities by omitting the traditional decoder structure, as our output is a scalar value rather than a sequence.
\begin{figure}[tbh]
	\centering
	\includegraphics[width=0.7\columnwidth]{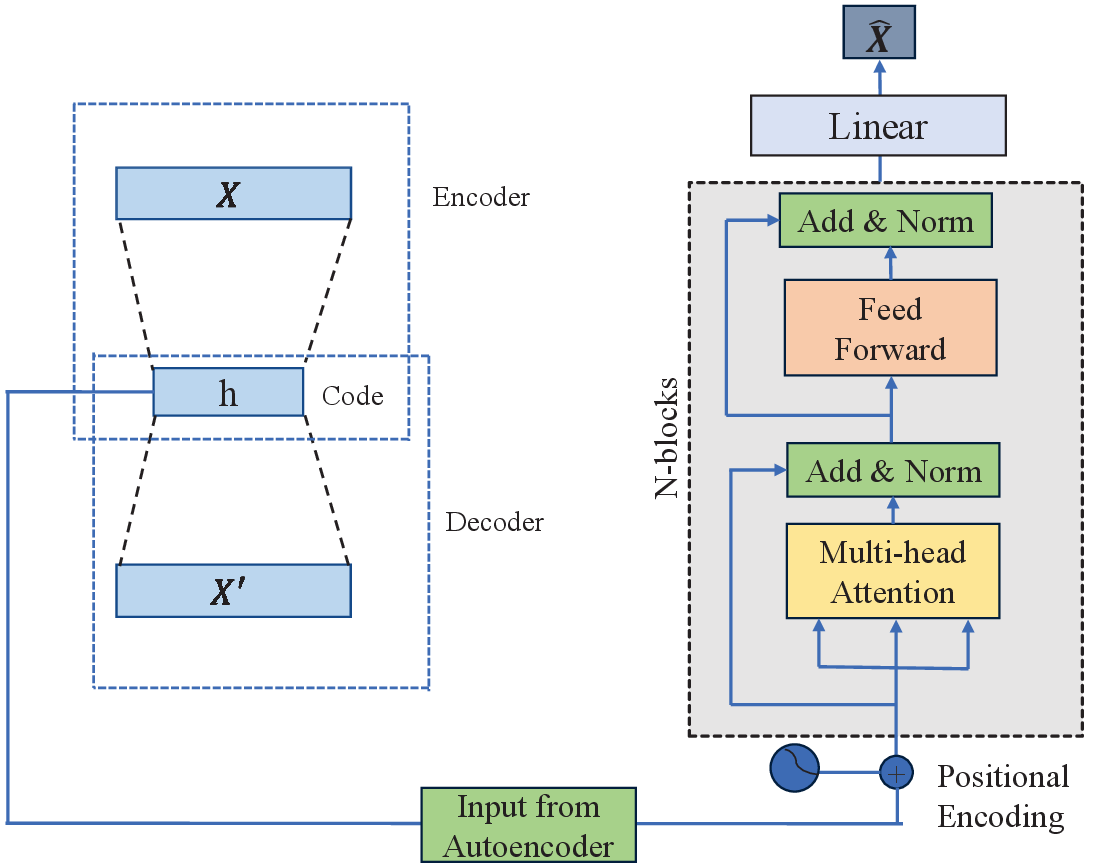}
	\caption{The proposed auto-encoder and transformer architecture for battery capacity prediction.}
	\label{fig3}
\end{figure}

As in Fig. 3, under the CC-CV protocol, the time it takes for the voltage to reach the upper threshold decreases as the charging cycles progress and the battery degrades. Also, the current will switch to the constant-voltage mode at an earlier time progressively, with the temperature reaching the maximum peak with less time. Note that for the CC-CV protocol, the duration of each charging cycle differs, resulting in different data length for each cycle. To address this issue, we use the subsampling technique to downsample the profiles of each cycle to be the same length prior to being passed to the transformer network. 


\begin{figure*}[t]
	\centering
	\includegraphics[width=0.8\textwidth]{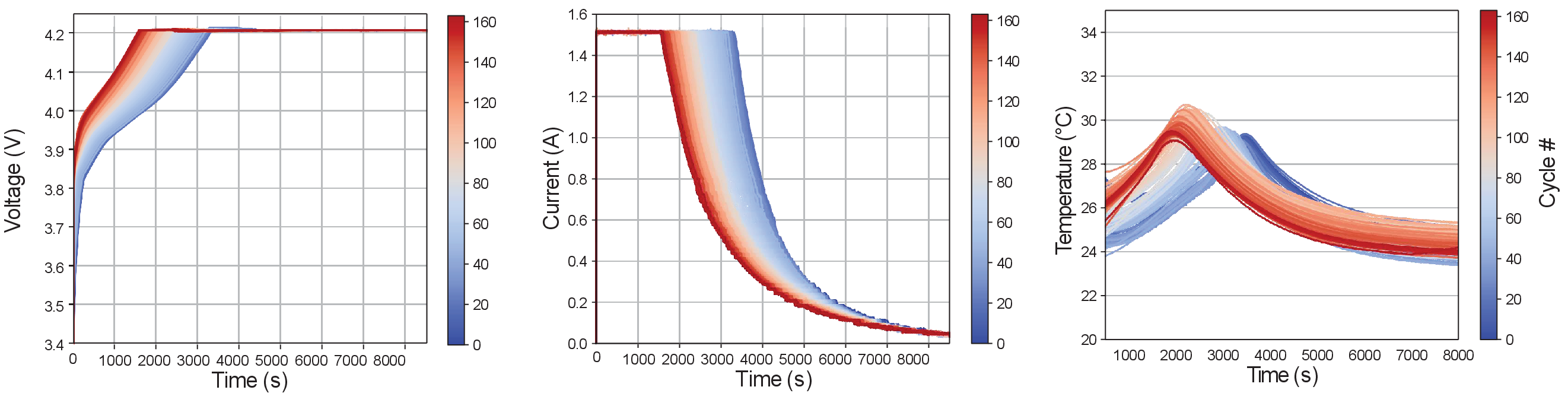}
	\vspace{-6 pt}
	\caption{Voltage, current and temperature profile for battery B0005.}
	\label{fig2}
\end{figure*}

\subsection{Architecture of the Proposed Transformer Network}
The structure of the proposed transformer network is shown in Fig. \ref{fig3}. For a generic charging cycle $t$, we define the down-sampled current, voltage, and temperature profiles as $I_{t}\in\mathbb{R}^{m}$, $V_{t}\in\mathbb{R}^{m}$, and $T_{t}\in\mathbb{R}^{m}$, respectively, with a scalar capacity value $c_{t}\in\mathbb{R}$. Also, we define the stacked multiple profiles as $X_t=[I_t, V_t, T_t, c_t]\in\mathbb{R}^{q}$, $q=3m+1\gg1$, which will be used to predict future capacities $c_{k}$, $k> t$. To capture the dynamics in the multivariate time-series, we consider a moving window of past $w$ cycles as the input sequence to the transformer: $\left\{X_{t-w+1},\ldots,X_{t}\right\}$. To tackle the potential high dimensions of $X_i$ at each cycle $i\in\left\{t-w+1,\ldots,t\right\}$, we first use an auto-encoder to reduce its dimension to a latent state $h_i\in\mathbb{R}^{l}$, $l\ll q$, with encoder and decoder respectively:
\begin{equation}
h_i=\mathcal{E}_{\theta_{1}}\left(X_{i}\right),~~X^{\prime}_i=\mathcal{D}_{\theta_1}\left(h_i\right),~~i=t-w+1,\ldots,t,
\end{equation}
where $X^{\prime}_i$ is the reconstructed profile from the decoder $\mathcal{D}_{\theta_1}(\cdot)$, and $\theta_1$ stands for the parameters of the auto-encoder network. We expect the extracted latent state $h_i$ to be representative of the core features of multiple profiles at cycle $i$. Then, the latent state sequence $\left\{h_{t-w+1},\ldots,h_{t}\right\}$ is fed into the transformer model as the input. The positional encoding as in \eqref{eq: posenc} is adopted to incorporate position information into the input. A number of transformer blocks, each block consisting of a multi-head attention mechanism and feed-forward network, is used to predict the next-step capacity value $c_{t+1}$. We define the series of transformer blocks as a mapping:
\begin{equation}
\hat{c}_{t+1}=\mathcal{T}_{\theta_{2}}\left(h_{t-w+1},\ldots,h_{t}\right),
\end{equation}
where $\theta_{2}$ is the collection of parameters of all blocks in the transformer architecture in Fig. \ref{fig3}, and $\hat{c}_{t+1}$ is the predicted next-cycle capacity value, which will be compared with the ground-truth capacity ${c}_{t+1}$. Ultimately, optimizing the parameters of the entire auto-encoder and transformer framework is formulated as:  
\begin{equation}
\min_{\theta}~~\sum_{t=w}^{n-1}\left(c_{t+1}-\hat{c}_{t+1}\right)^2 + \lambda \sum_{t=w}^{n-1}\sum_{i=t-w+1}^{t}\left(X^{\prime}_{i}-X_{i}\right)^2,
\end{equation}
where $n$ is the available number of moving windows created from training data, $\theta=\{\theta_1,\theta_2\}$ stacks all network parameters, and $\lambda\ge0$ is the weight controlling the relative importance between two objectives: transformer prediction error and auto-encoder reconstruction error. 

\section{Case Studies and Discussions}
\label{sec: Results and Discussion}
In this section, we employ benchmark NASA Group 1 and UofM battery degradation datasets to validate the proposed methods in previous sections.

\subsection{Setup of Case Studies}
For these case studies, the hardware configurations include an Intel(R) Xeon(R) W-2265 CPU @ 3.50 GHz processor, 64.0 GB RAM, and an NVIDIA RTX A4000 graphics card. The software environment comprises the PyTorch operating on Python version 3.9. For each battery group, we predict the last 60 cycles of each cell while using the preceding cycles and data from the other 3 cells as the training data. We use the voltage, current, temperature, and the capacity values of each cycle. However, since each cycle has a different length of voltage, current, and temperature profiles, down-sampling is used to divide the profiles of each cycle to a desired dimension for the transformer. Specifically, we down-sample the profiles into $q=48$ dimensions, consisting of 16 voltage, 16 temperature, 15 current points, and one capacity value for each charging cycle. 
\begin{figure}[tbh]
	\centering
	\includegraphics[width = \columnwidth]{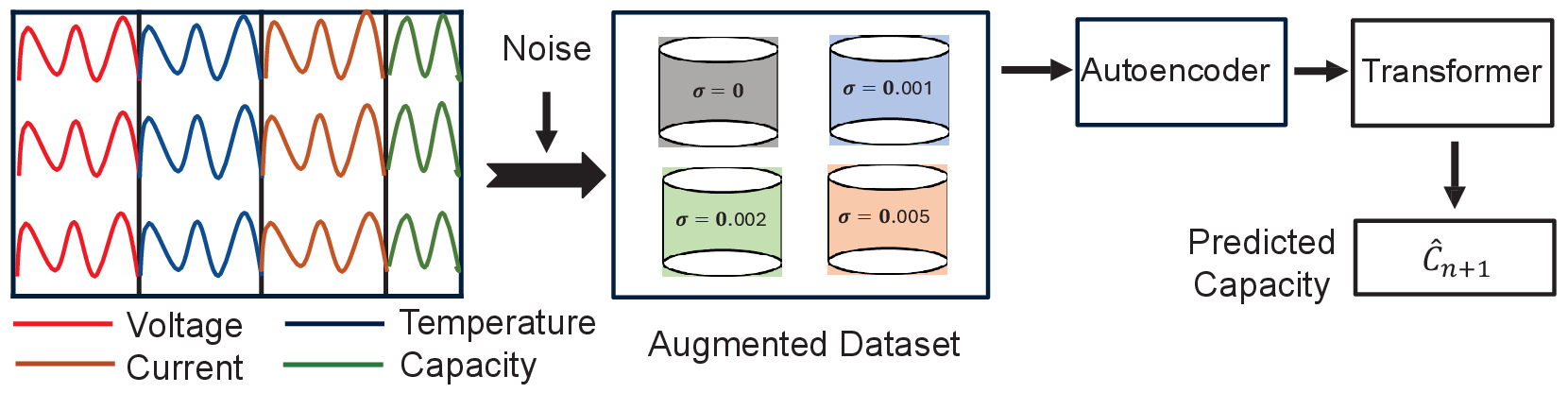}
	\caption{Logic flow of the case studies.}
	\label{fig4}
\end{figure}

To tackle the data scarcity, we use data augmentation by creating 3 additional artificial datasets, and each of them is obtained by adding white noise of a different level ($\sigma=0.001, 0.002,0.005$) to all the profiles (see Fig. \ref{fig4}). We observed that a large noise level resulted in decreased model performance. This decline is due to the fact that adding large white noise may destroy the inherent dynamics of the original data. For each dataset, a sliding window of $w=16$ cycles is created on the temperature, voltage, and current profiles. For the autoencoder, the dimension of latent state is chosen as $l=8$.



The prediction of future capacities is conducted in a rolling manner, where the predicted $\hat{c}_{n+1}$ ($n$ is the current cycle index) will be incorporated to construct $X_{n+1}$, and $I_{n+1}$, $V_{n+1}$, $T_{n+1}$ are assumed to be known, as in practice these profiles are easily acquired from the battery management system, in contrast to the capacity value. 



\begin{figure*}[!h]
	\centering
	\includegraphics[width=\linewidth]{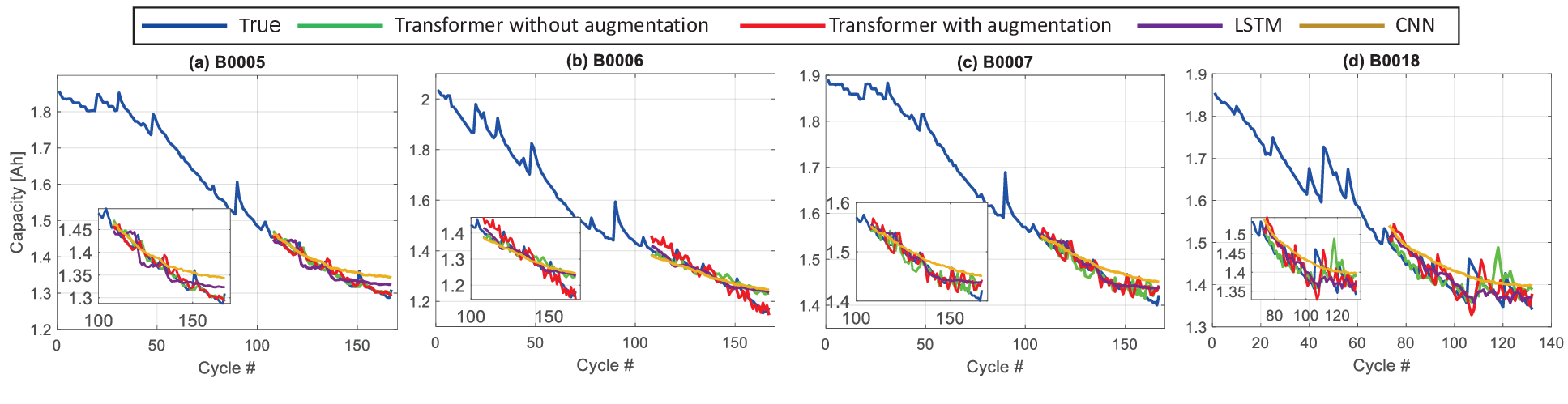}
	\vspace{-20pt}
	\caption{Predicted capacity vs. ground-truth values for NASA Group 1 dataset.}
	\label{fig5}
\end{figure*}

\begin{figure*}[!h]
	\centering
	\includegraphics[width=\linewidth]{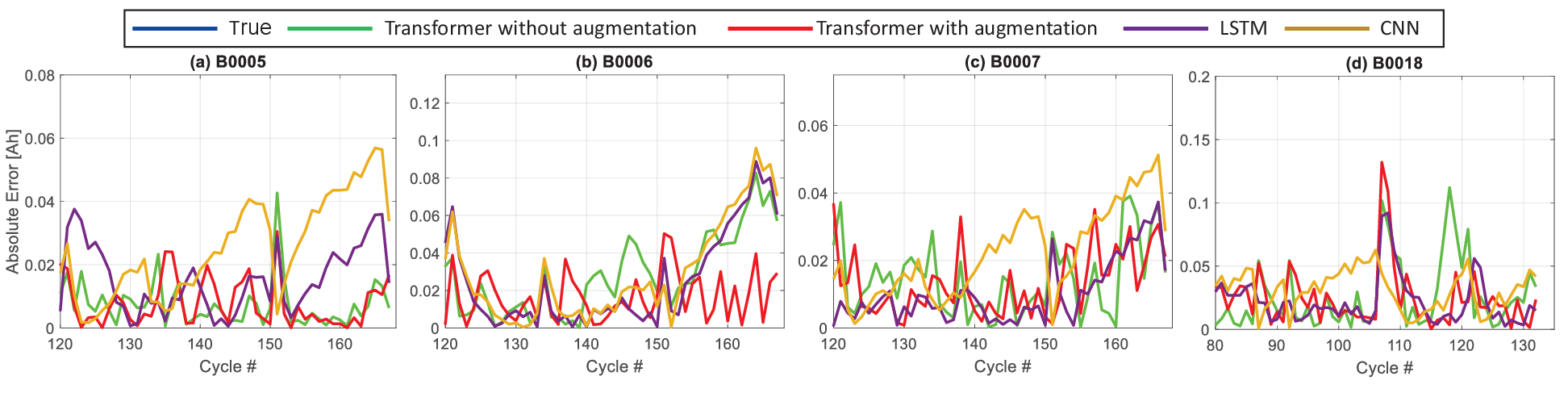}
	\vspace{-20pt}
	\caption{Absolute prediction error of capacities for NASA Group 1 dataset.}
	\label{fig6}
\end{figure*}

\begin{figure*}[!h]
	\centering
	\includegraphics[width=\linewidth]{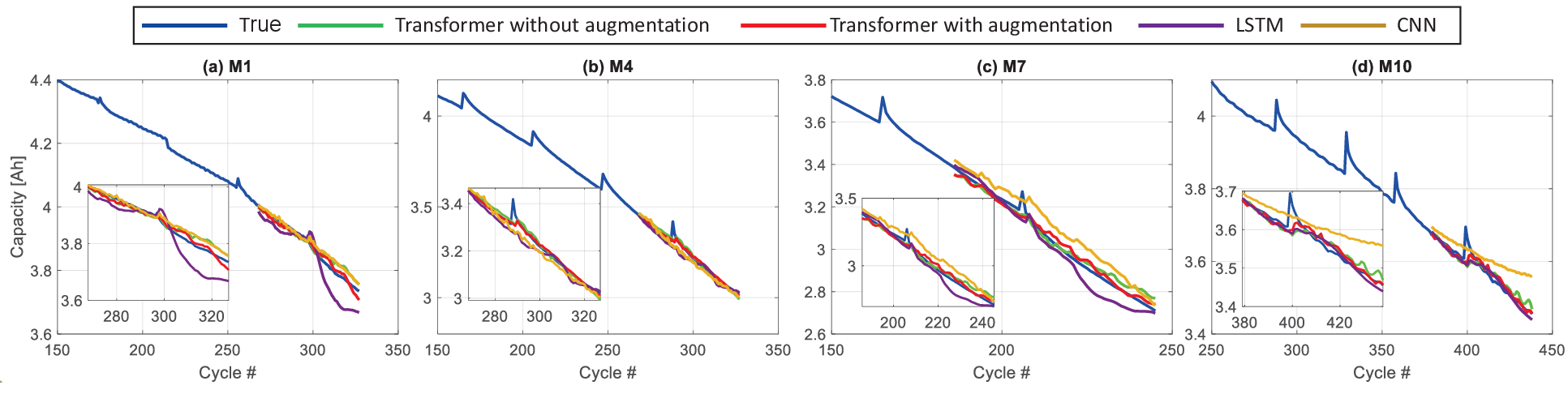}
	\vspace{-20pt}
	\caption{Predicted capacity vs. ground-truth values for UofM dataset.}
	\label{fig7}
\end{figure*}

\begin{figure*}[!h]
	\centering
	\includegraphics[width=\linewidth]{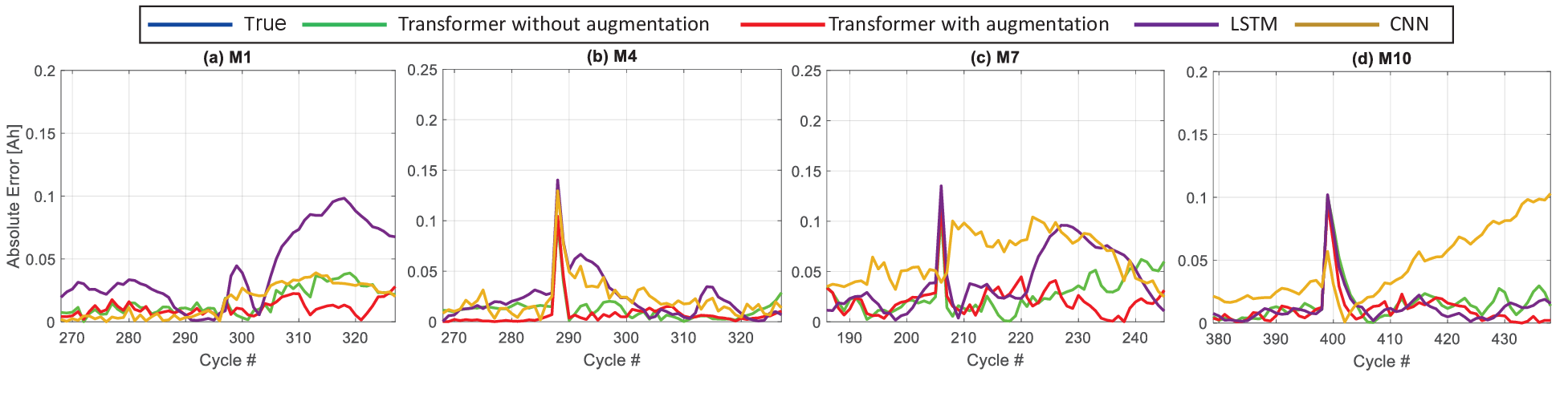}
	\vspace{-20pt}
	\caption{Absolute prediction error of capacities for UofM dataset.}
	\label{fig8}
\end{figure*}

\subsection{Experimental Result and Analysis:}
To assess the efficiency of the model, the mean absolute error (MAE) and root mean square error (RMSE) metrics below are used as performance indicators:
\begin{equation}
\text{MAE} = \frac{1}{p}\sum_{i=1}^{p}\left| (y_i - \hat{y}_i)\right|, 	\text{RMSE} = \sqrt{\frac{1}{p} \sum_{i=1}^{p} (y_i - \hat{y}_i)^2}
\end{equation}
where $p$ is the number of capacities to be predicted, $y_i$ is the true capacity while $\hat{y}_i$ is the predicted capacity.

Figs. \ref{fig5} and \ref{fig7} show the predicted capacity values versus the ground truth values, while Figs. \ref{fig6} and \ref{fig8} show the absolute error plot for the NASA Group 1 and UofM datasets, respectively. Four methods are compared: the transformer with data augmentation (our proposed method), transformer without data augmentation, LSTM, and CNN. Notably, the tests for CNN and LSTM were performed using the augmented data. Tables 1-2 illustrate the specific metrics for the prediction errors from these methods. From these results, it is observed that the proposed transformer with data augmentation method consistently outperforms the other approaches. Moreover, overall, the transformer methods excel CNN and LSTM largely, which validates the benefits of using transformer to extract more useful (e.g., long-term) patterns for better capacity prediction. 

\begin{table}[!h]
	\centering
	\caption{Capacity prediction for NASA data.}
	\label{tab:tab1}
	{\scriptsize 
		\begin{tabularx}{0.9\columnwidth}{c c c c c c }
			\hline
			Battery & Criteria  & CNN & LSTM & Transf. &  \parbox[t]{1cm}{Transf. \\ w. aug.}\\
			\hline
			\multirow{2}{*}{B0005} & MAE & 0.0223 & 0.0156 & 0.0087 & \textbf{0.0074} \\
			& RMSE  & 0.2093 & 0.1466 &0.0098& \textbf{0.0078} \\
			\hline
			\multirow{2}{*}{B0006} & MAE & 0.0258 & \textbf{0.0223} & 0.0228 & \textbf{0.0223} \\
			& RMSE  & 0.2817 & 0.2534 & 0.0029 & \textbf{0.0026} \\
			\hline
			\multirow{2}{*}{B0007} & MAE & 0.0193 & \textbf{0.0111} & 0.0139 & 0.0131 \\
			& RMSE  & 0.1810 & 0.1105 & 0.0173 & \textbf{0.0157} \\
			\hline
			\multirow{2}{*}{B0018} & MAE & 0.0308 & \textbf{0.0220} & 0.0238 & 0.0235  \\
			& RMSE  & 0.2488 & 0.2185 & 0.0320 & \textbf{0.0275} \\
			\hline
		\end{tabularx}
	}
\end{table}

\begin{table}[!h]
	\centering
	\caption{Capacity prediction for UofM data.}
	\label{tab:tab2}
	{\scriptsize 
		\begin{tabularx}{0.9\columnwidth}{c c c c c c }
			\hline
			Battery & Criteria  & CNN & LSTM & Transf. & \parbox[t]{1cm}{Transf. \\ w. aug.}\\
			\hline
			\multirow{2}{*}{M1} & MAE & 0.0163 & 0.0424 & 0.0165 & \textbf{0.0108} \\
			& RMSE & 0.1202 & 0.1219 & 0.0197 & \textbf{0.0123} \\
			\hline
			\multirow{2}{*}{M4} & MAE & 0.0223 & 0.0236 & 0.0125 & \textbf{0.0072} \\
			& RMSE & 0.2568 & 0.2967 & 0.0193 & \textbf{0.0159} \\
			\hline
			\multirow{2}{*}{M7} & MAE & 0.0642 & 0.0417 & 0.0262 & \textbf{0.0212} \\
			& RMSE & 0.4429 & 0.4565 & 0.0322& \textbf{0.0268} \\
			\hline
			\multirow{2}{*}{M10} & MAE & 0.0448 & 0.0129 & 0.0158 & \textbf{0.0110}  \\
			& RMSE & 0.2857 & 0.1624 & 0.0229 & \textbf{0.0183} \\
			\hline
		\end{tabularx}
	}
\end{table}

\section{Conclusion}
\label{sec: Conclusion}
In this work we proposed a framework integrating auto-encoder and transformer network for battery capacity prediction. To address the data scarcity issue, we use data augmentation to enrich the dataset by adding random noise to mimic the signal fluctuations and sensor inaccuracies occurring in real-life situations. We further use the well-known benchmark NASA Group 1 and UofM datasets for validating the effectiveness of the proposed methods. Simulation results show that transformer networks consistently outperform the other deep learning methods such as LSTM and CNN in giving accurate battery capacity predictions. Moreover, with data augmentation, the performance of transformer networks can be further improved when trained with a larger data size. In this work, the variance of the random noise used to enrich the dataset is kept small, so that the noise will not distort the inherent dynamic behaviors of battery signals. However, since random noise cannot represent the dynamics of the original signal, one future direction is to exploit other data augmentation techniques to increase the data size and richness to better address the data scarcity issue. Additionally, we hope to explore datasets collected under different temperature conditions to further enhance the robustness and applicability of our model. 

\section{Acknowledgment}
We acknowledge the support from NSF Grant 2340194.


\bibliography{References}  

\begin{thebibliography}{25}
\providecommand{\natexlab}[1]{#1}
\providecommand{\url}[1]{\texttt{#1}}
\providecommand{\urlprefix}{URL }
\expandafter\ifx\csname urlstyle\endcsname\relax
  \providecommand{\doi}[1]{doi:\discretionary{}{}{}#1}\else
  \providecommand{\doi}{doi:\discretionary{}{}{}\begingroup
  \urlstyle{rm}\Url}\fi

\bibitem[{Barcellona et~al.(2022)Barcellona, Colnago, Dotelli, Latorrata, and
  Piegari}]{barcellona2022aging}
Barcellona, S., Colnago, S., Dotelli, G., Latorrata, S., and Piegari, L.
  (2022).
\newblock Aging effect on the variation of li-ion battery resistance as
  function of temperature and state of charge.
\newblock \emph{Journal of Energy Storage}, 50, 104658.

\bibitem[{Catelani et~al.(2021)Catelani, Ciani, Fantacci, Patrizi, and
  Picano}]{catelani2021remaining}
Catelani, M., Ciani, L., Fantacci, R., Patrizi, G., and Picano, B. (2021).
\newblock Remaining useful life estimation for prognostics of lithium-ion
  batteries based on recurrent neural network.
\newblock \emph{IEEE Transactions on Instrumentation and Measurement}, 70,
  1--11.

\bibitem[{Chen et~al.(2022)Chen, Hong, and Zhou}]{chen2022transformer}
Chen, D., Hong, W., and Zhou, X. (2022).
\newblock Transformer network for remaining useful life prediction of
  lithium-ion batteries.
\newblock \emph{IEEE Access}, 10, 19621--19628.

\bibitem[{Choi et~al.(2019)Choi, Ryu, Park, and Kim}]{choi2019machine}
Choi, Y., Ryu, S., Park, K., and Kim, H. (2019).
\newblock Machine learning-based lithium-ion battery capacity estimation
  exploiting multi-channel charging profiles.
\newblock \emph{IEEE Access}, 7, 75143--75152.

\bibitem[{Fan et~al.(2022)Fan, Wu, Chen, Jiang, Huang, and Chen}]{fan2022data}
Fan, Y., Wu, H., Chen, W., Jiang, Z., Huang, X., and Chen, S.Z. (2022).
\newblock A data augmentation method to optimize neural networks for predicting
  soh of lithium batteries.
\newblock In \emph{Journal of Physics: Conference Series}, volume 2203, 012034.
  IOP Publishing.

\bibitem[{Gu et~al.(2023)Gu, See, Li, Shan, Wang, Zhao, Lim, and
  Zhang}]{gu2023novel}
Gu, X., See, K., Li, P., Shan, K., Wang, Y., Zhao, L., Lim, K.C., and Zhang, N.
  (2023).
\newblock A novel state-of-health estimation for the lithium-ion battery using
  a convolutional neural network and transformer model.
\newblock \emph{Energy}, 262, 125501.

\bibitem[{Guo et~al.(2022)Guo, Sun, Vilsen, Meng, and Stroe}]{guo2022review}
Guo, W., Sun, Z., Vilsen, S.B., Meng, J., and Stroe, D.I. (2022).
\newblock Review of “grey box” lifetime modeling for lithium-ion battery:
  Combining physics and data-driven methods.
\newblock \emph{Journal of Energy Storage}, 56, 105992.

\bibitem[{Hannan et~al.(2021)Hannan, How, Lipu, Mansor, Ker, Dong, Sahari,
  Tiong, Muttaqi, Mahlia et~al.}]{hannan2021deep}
Hannan, M.A., How, D.N., Lipu, M.H., Mansor, M., Ker, P.J., Dong, Z., Sahari,
  K., Tiong, S.K., Muttaqi, K.M., Mahlia, T.I., et~al. (2021).
\newblock Deep learning approach towards accurate state of charge estimation
  for lithium-ion batteries using self-supervised transformer model.
\newblock \emph{Scientific Reports}, 11(1), 19541.

\bibitem[{Horiba(2014)}]{horiba2014lithium}
Horiba, T. (2014).
\newblock Lithium-ion battery systems.
\newblock \emph{Proceedings of the IEEE}, 102(6), 939--950.

\bibitem[{Hussein(2014)}]{hussein2014capacity}
Hussein, A.A. (2014).
\newblock Capacity fade estimation in electric vehicle li-ion batteries using
  artificial neural networks.
\newblock \emph{IEEE Transactions on Industry Applications}, 51(3), 2321--2330.

\bibitem[{Lee et~al.(2023)Lee, Kwon, and Lee}]{lee2023convolutional}
Lee, G., Kwon, D., and Lee, C. (2023).
\newblock {A convolutional neural network model for SOH estimation of Li-ion
  batteries with physical interpretability}.
\newblock \emph{Mechanical Systems and Signal Processing}, 188, 110004.

\bibitem[{Liu et~al.(2021)Liu, Ren, Zhang, Sun, and Zou}]{liu2021rethinking}
Liu, F., Ren, X., Zhang, Z., Sun, X., and Zou, Y. (2021).
\newblock Rethinking skip connection with layer normalization in transformers
  and resnets.
\newblock \emph{arXiv preprint arXiv:2105.07205}.

\bibitem[{Lucaferri et~al.(2023)Lucaferri, Quercio, Laudani, and
  Riganti~Fulginei}]{lucaferri2023review}
Lucaferri, V., Quercio, M., Laudani, A., and Riganti~Fulginei, F. (2023).
\newblock A review on battery model-based and data-driven methods for battery
  management systems.
\newblock \emph{Energies}, 16(23), 7807.

\bibitem[{Miao et~al.(2019)Miao, Hynan, Von~Jouanne, and
  Yokochi}]{miao2019current}
Miao, Y., Hynan, P., Von~Jouanne, A., and Yokochi, A. (2019).
\newblock Current li-ion battery technologies in electric vehicles and
  opportunities for advancements.
\newblock \emph{Energies}, 12(6), 1074.

\bibitem[{Mohtat et~al.(2021)Mohtat, Lee, Siegel, and
  Stefanopoulou}]{mohtat2021reversible}
Mohtat, P., Lee, S., Siegel, J.B., and Stefanopoulou, A.G. (2021).
\newblock Reversible and irreversible expansion of lithium-ion batteries under
  a wide range of stress factors.
\newblock \emph{Journal of The Electrochemical Society}, 168(10), 100520.

\bibitem[{Morita et~al.(2021)Morita, Saito, Yoshioka, and
  Shiratori}]{morita2021estimation}
Morita, Y., Saito, Y., Yoshioka, T., and Shiratori, T. (2021).
\newblock Estimation of recoverable resources used in lithium-ion batteries
  from portable electronic devices in japan.
\newblock \emph{Resources, Conservation and Recycling}, 175, 105884.

\bibitem[{Mumuni and Mumuni(2022)}]{mumuni2022data}
Mumuni, A. and Mumuni, F. (2022).
\newblock Data augmentation: A comprehensive survey of modern approaches.
\newblock \emph{Array}, 16, 100258.

\bibitem[{Park et~al.(2020)Park, Choi, Choi, Ryu, and Kim}]{park2020lstm}
Park, K., Choi, Y., Choi, W.J., Ryu, H.Y., and Kim, H. (2020).
\newblock {LSTM-based battery remaining useful life prediction with
  multi-channel charging profiles}.
\newblock \emph{IEEE Access}, 8, 20786--20798.

\bibitem[{Saha and Goebel(2007)}]{saha2007battery}
Saha, B. and Goebel, K. (2007).
\newblock Battery data set.
\newblock \emph{NASA AMES Prognostics Data Repository}.

\bibitem[{Vaswani et~al.(2017)Vaswani, Shazeer, Parmar, Uszkoreit, Jones,
  Gomez, Kaiser, and Polosukhin}]{vaswani2017attention}
Vaswani, A., Shazeer, N., Parmar, N., Uszkoreit, J., Jones, L., Gomez, A.N.,
  Kaiser, {\L}., and Polosukhin, I. (2017).
\newblock Attention is all you need.
\newblock \emph{Advances in Neural Information Processing Systems}, 30.

\bibitem[{Wolf et~al.(2020)Wolf, Debut, Sanh, Chaumond, Delangue, Moi, Cistac,
  Rault, Louf, Funtowicz et~al.}]{wolf2020transformers}
Wolf, T., Debut, L., Sanh, V., Chaumond, J., Delangue, C., Moi, A., Cistac, P.,
  Rault, T., Louf, R., Funtowicz, M., et~al. (2020).
\newblock Transformers: State-of-the-art natural language processing.
\newblock In \emph{Proceedings of the 2020 Conference on Empirical Methods in
  Natural Language Processing: System Demonstrations}, 38--45.

\bibitem[{Wu et~al.(2016)Wu, Fu, and Guan}]{wu2016review}
Wu, L., Fu, X., and Guan, Y. (2016).
\newblock Review of the remaining useful life prognostics of vehicle
  lithium-ion batteries using data-driven methodologies.
\newblock \emph{Applied Sciences}, 6(6), 166.

\bibitem[{Yao et~al.(2021)Yao, Xu, Tang, Zhou, Hou, Xiao, and
  Fu}]{yao2021review}
Yao, L., Xu, S., Tang, A., Zhou, F., Hou, J., Xiao, Y., and Fu, Z. (2021).
\newblock A review of lithium-ion battery state of health estimation and
  prediction methods.
\newblock \emph{World Electric Vehicle Journal}, 12(3), 113.

\bibitem[{Zhang et~al.(2023)Zhang, Yang, Du, Sun, Li, Wang, and
  Wang}]{zhang2023review}
Zhang, M., Yang, D., Du, J., Sun, H., Li, L., Wang, L., and Wang, K. (2023).
\newblock {A review of SOH prediction of Li-ion batteries based on data-driven
  algorithms}.
\newblock \emph{Energies}, 16(7), 3167.

\bibitem[{Zhang et~al.(2018)Zhang, Xiong, He, and Pecht}]{zhang2018long}
Zhang, Y., Xiong, R., He, H., and Pecht, M.G. (2018).
\newblock Long short-term memory recurrent neural network for remaining useful
  life prediction of lithium-ion batteries.
\newblock \emph{IEEE Transactions on Vehicular Technology}, 67(7), 5695--5705.

\end{thebibliography}

\end{document}